\pdfoutput=1

\documentclass[11pt]{article}

\usepackage[]{acl}

\usepackage{times}
\usepackage{latexsym}

\usepackage[T1]{fontenc}

\usepackage[utf8]{inputenc}

\usepackage{microtype}

\usepackage{inconsolata}
\usepackage[utf8]{inputenc} %
\usepackage[T1]{fontenc}    %
\usepackage{microtype,inconsolata}
\usepackage{times,latexsym}
\usepackage{graphicx} \graphicspath{{figures/}}
\usepackage{amsmath,amssymb,mathabx,mathtools,amsthm,nicefrac}
\usepackage[linesnumbered,ruled,vlined]{algorithm2e}
\usepackage{acronym}
\usepackage{enumitem}
\usepackage{balance}
\usepackage{xspace}
\usepackage{setspace}
\usepackage[capitalise,noabbrev,nameinlink]{cleveref}
\usepackage{booktabs,tabularx,colortbl,multirow,multicol,array,makecell,tabularray}
\usepackage{overpic,wrapfig}
\usepackage[misc]{ifsym}
\usepackage{pifont}

\usepackage{times}
\usepackage{latexsym}
\usepackage{xcolor}
\usepackage{soul}
\usepackage{amsmath}
\usepackage{booktabs}
\usepackage{xcolor}
\usepackage{url}
\usepackage[T1]{fontenc}
\usepackage{longtable}
\usepackage{graphicx}
\usepackage{float}
\usepackage{multirow}
\usepackage{color}
\usepackage{wrapfig}
\usepackage{booktabs}
\usepackage{amssymb}
\usepackage{subfigure}
\usepackage{listings}
\usepackage{pifont}
\usepackage{wasysym}
\usepackage{utfsym}
\usepackage{fontawesome}
\usepackage{makecell}

\usepackage[T1]{fontenc}

\usepackage[utf8]{inputenc}

\usepackage{microtype}

\usepackage{inconsolata}

\makeatletter
\DeclareRobustCommand\onedot{\futurelet\@let@token\@onedot}
\def\@onedot{\ifx\@let@token.\else.\null\fi\xspace}

\makeatother

\makeatletter
\renewcommand{\paragraph}{%
  \@startsection{paragraph}{4}%
  {\z@}{0ex \@plus 0ex \@minus 0ex}{-1em}%
  {\hskip\parindent\normalfont\normalsize\bfseries}%
}
\makeatother

\crefname{algorithm}{Alg.}{Algs.}
\Crefname{algocf}{Algorithm}{Algorithms}
\crefname{section}{Sec.}{Secs.}
\Crefname{section}{Section}{Sections}
\crefname{table}{Tab.}{Tabs.}
\Crefname{table}{Table}{Tables}
\crefname{figure}{Fig.}{Figs.}
\Crefname{figure}{Figure}{Figures}
\crefname{equation}{Eq.}{Eqs.}
\Crefname{equation}{Equation}{Equations}
\crefname{appendix}{Appx.}{Appxs.}
\Crefname{appendix}{Appendix}{Appendices}

\definecolor{gblue}{HTML}{4285F4}
\definecolor{gred}{HTML}{DB4437}
\definecolor{ggreen}{HTML}{0F9D58}

\usepackage{fancyvrb}
\usepackage{fvextra}
\usepackage{csquotes}
\usepackage{scalerel}

\hypersetup{
  citecolor=gray, %
  linkcolor=magenta,
  anchorcolor=blue,
}

\definecolor{mygray}{gray}{.92}
\definecolor{emphypurple}{rgb}{0.302, 0.055, 0.659}
\newcommand{\tabemph}[1]{\cellcolor{mygray!100}\textcolor{black!100!mygray}{\textbf{#1}}}
\definecolor{highlightgreen}{HTML}{009901}
\definecolor{highlightred}{HTML}{FD6864}

\acrodef{16pf}[16PF]{Sixteen Personality Factors}
\acrodef{big5}[Big Five]{Big Five Personality Factors}
\acrodef{ctg}[CTG]{Controllable Text Generation}
\acrodef{nlp}[NLP]{Natural Language Processing}
\acrodef{mpi}[MPI]{Machine Personality Inventory}
\acrodef{prolific}[Prolific]{Prolific Academic Ltd}
\acrodef{ipip}[IPIP]{International Personality Item Pool}
\acrodef{mbti}[MBTI]{Myers-Briggs Type Indicator}
\acrodef{mmpi}[MMPI]{Minnesota Multiphasic Personality Inventory}
\acrodef{llm}[LLM]{Large Language Model}
\acrodef{method}[\textsc{P}$^2$]{\textsc{Personality Prompting}}

\title{ControlLM: Crafting Diverse Personalities for Language Models}

\author{Yixuan Weng$^1$, Shizhu He$^{1,2}$, Kang Liu$^{1,2}$, Shengping Liu$^{3}$, Jun Zhao$^{1,2}$ \\ 
	$^1$ The Laboratory of Cognition and Decision Intelligence for Complex Systems, IA, CAS \\
	$^2$ School of Artificial Intelligence, University of Chinese Academy of Sciences\\
  $^3$Unisound, Beijing, China \\
\texttt{{wengsyx@gmail.com}, {\{shizhu.he, kliu, jzhao\}@nlpr.ia.ac.cn}} \\
{\textcolor[RGB]{228,0,127}{https://github.com/wengsyx/ControlLM}}
}

\begin{document}
\maketitle

\begin{abstract}

As language models continue to scale in size and capability, they display an array of emerging behaviors, both beneficial and concerning. This heightens the need to control model behaviors. We hope to be able to control the personality traits of language models at the inference-time so as to have various character features, on top of which the requirements of different types of tasks can be met. Personality is a higher-level and more abstract behavioral representation for language models. We introduce ControlLM, which leverages differential activation patterns, derived from contrasting behavioral prompts in the model's latent space, to influence the model's personality traits at inference. This approach allows for the precise, real-time adjustment of model behavior. First, we demonstrate ControlLM's capacity to elicit diverse persona behaviors without any training, while precision control allows personality traits to closely match average human values. Subsequently, we showcase improved reasoning and question answering through selective amplification of beneficial attributes like conscientiousness and friendliness. We hope that this work will inspire research on controlling human-like behaviors of language models and provide insights for future research. Our code is publicly
available at: \url{https://github.com/wengsyx/ControlLM}.

\end{abstract}

\section{Introduction}

\begin{figure}[t]
\begin{center}
\includegraphics[width=0.48\textwidth]{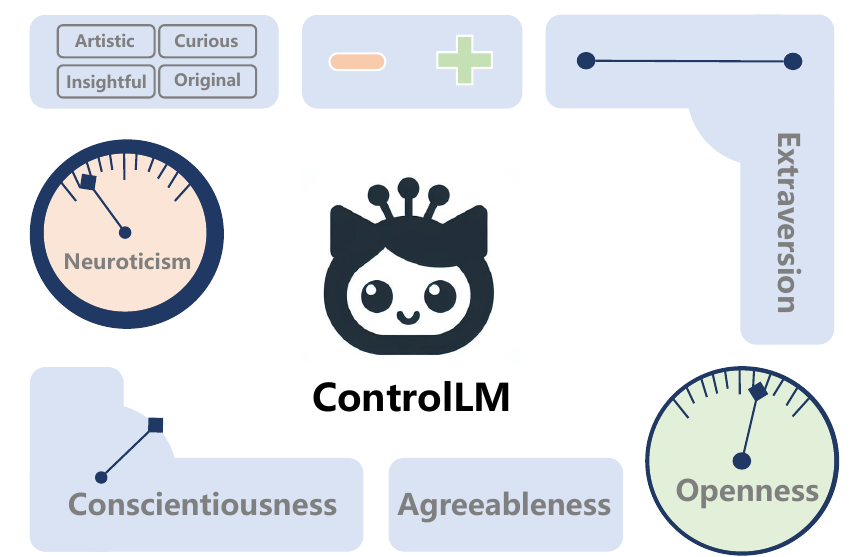}
\vspace{-0.65cm} 
\end{center}
\caption{ControlLM adjusts LM's outputs by modulating key personality traits, enabling the real-time craftion of diverse personas with fine-grained behavioral control.}
\vspace{-0.2cm} 
\label{figure:1}
\end{figure}






As Large Language Models (LLMs) evolve, their capabilities exhibit increasingly complex behaviors and personalities \cite{zhou2022large,wang2022self,chen2023teaching}. Studies show LLMs like GPT-4 \cite{openai2023gpt4} can manifest distinct human-like personas \cite{Binz_2023,aher2023using,dasgupta2023language}, posing challenges in aligning their actions with human intent \cite{jiang2023evaluating}. This rapid convergence toward human-like behavior begs the question: how can we prevent LLMs from exhibiting harmful behaviors similar to those observed in humans with abnormal psychological states? We propose fine-grained control of LLM personalities and behaviors, enabling the shaping of their actions and priorities to uphold desired values and principles.

There are differences between LMs' hidden representations and output contents. \citet{saunders2022selfcritiquing} coined the term "Generation-Discrimination gap" (G-D gap) for this phenomenon. Prior research indicates that LMs internally "know" more than they "express" in their output, suggesting the potential for modulating those hidden states to refine output \cite{burns2022discovering,li2023inferencetime}.

We introduce ControlLM to enable fine-grained influence over language model personalities without expensive retraining. Our approach involves shifting activation patterns along interpretable directions representing salient attributes. Specifically, we first obtain a small set of opposing behavioral prompts for personality and extract their difference vectors within the model's latent space, inspired by recent work on activation addition \cite{rimsky2023steering,jorgensen2023improving}. These differential activations become control directions linking inputs to associated personalities. At inference time, we can select any combination of directions and modulate scaling factors to elicit desired model behaviors across self-regressive generations.


ControlLM provides granularity and specificity in modulating model behaviors through its ability to manipulate personality traits at inference time. This fine-adjustment capability enables LLMs to adapt their responses according to the nuanced demands of various tasks. For instance, one could accentuate an LLM’s sense of responsibility to bolster its performance in reasoning tasks \citep{suzgun2022challenging}. Moreover, ControlLM tackles a crucial challenge: mitigating harmful model behaviors, particularly the sycophancy tendency observed in RLHF-trained \cite{LongOuyang2022TrainingLM} models. This tendency compromises the truthfulness of model outputs and erodes trust in their applications \citep{yang2023alignment, bai2022training}. Through ControlLM, we can enforce fine-grained alignment and instill an LLM with tailored traits such as increased truthfulness \citep{Min2023FActScoreFA}. Furthermore, by enabling LLMs to assume diverse personas, ControlLM contributes to the crafting of a heterogeneous synthetic society, an innovative approach with vast potential for studying social dynamics and human-computer interaction within artificially constructed environments \citep{park2023generative,chen2023large}. ControlLM offers nuanced and context-specific control, paving the way for LLMs that are both versatile and value-aligned. Our contributions are as follows:



\begin{itemize}
    \item Empower language models with fine-grained control over specific personality roles by dynamically adjusting activation representations to align with desired personality traits. (See in {\textcolor[RGB]{228,0,127} {Section}} \ref{E:1}.)
    \item Making language models more diligent and reliable during complex reasoning processes, e.g. increasing conscientiousness. (See in {\textcolor[RGB]{228,0,127} {Section}} \ref{E:2}.)
    \item Enhance the general conversational capacity of LMs by tempering attributes like openness, while fostering more agreeable characteristics such as friendliness. (See in {\textcolor[RGB]{228,0,127} {Section}} \ref{E:3} and {\textcolor[RGB]{228,0,127} {Section}} \ref{E:4}.)
    \item Address and mitigate specific maladaptive behaviors, such as a propensity for sycphancy, by implementing timely, targeted control measures to maintain the integrity of the original model's capabilities without retraining. (See in {\textcolor[RGB]{228,0,127} {Section}} \ref{E:4}.)
    \item Introducing the AutoControlActivate toolkit, which can obtain the activation vector of a specific personality within one minute. This increases the generality ability of ControlLM. (See in {\textcolor[RGB]{228,0,127} {Section}} \ref{M:1})

\end{itemize}

\section{Related Work}




\noindent\textbf{Personas and agents in LLMs.} Recent advancements in LMs have demonstrated their ability to capture and replicate complex human-like behaviors \cite{andreas-2022-language,durmus2023measuring,shanahan2023role} when they attain a sufficient scale in training data and parameters \cite{wei2022emergent}. In some case studies and standardized assessments of language models simulating social sciences \cite{Binz_2023,jiang2023mewl,jiang2023evaluating}, it has been found that language models themselves have certain personality traits to some degree. Traditionally, personality describes human behaviour patterns \cite{corsini1994encyclopedia,weinberg2023foundations}. There have been instances where modeling language models to adopt roles, such as that of a professor, increases their propensity for truthful outputs \cite{zhou2022large}. 
While prompting can influence LLM behavior, precisely capturing and characterizing their personality and behavior through prompts is complex and unreliable, making large-scale personalized control impractical.

\noindent\textbf{LLMs' personality control.} A series of early studies focused primarily on controlled text generation \cite{zhang2023survey}, modifying sentences to achieve textual attributes while retaining irrelevant text content \cite{hu2017toward,yi2018automatic,li2018delete,dathathri2019plug}. Additionally, methods for Intervening on weight \cite{ranzato2015sequence,meng2022locating,liu2024tuning}, Intervening at decoding \cite{grover2019bias}, and Intervening on the prompt \cite{shin2020autoprompt,zhou2022steering} can also influence language model outputs. Later, \citet{turner2023activation}’s study found that by modifying language models activations during the forward-stage, language models' styles could be controlled – this is known as activation engineering (or representation engineering \cite{zou2023representation}). Building on existing research, our work aims to achieve nuanced control over LMs by manipulating higher-level personality representations. This fine-grained approach promises to infuse models with diverse character roles that embody different value systems, all while requiring minimal computational resources.

\noindent\textbf{LLMs’ downstream task.} The complex reasoning abilities and humanized question answering capabilities of LLMs have attracted widespread attention \cite{sun2023aligning,cui2023ultrafeedback,yao2023instructions}. Chain-of-thought (CoT) \cite{akyurek2022learning,weichain} is one of the most classic methods, which enables language models to generate explanatory processes before outputting answers through in-context learning \cite{XuezhiWangSelfConsistencyIC,dong2022survey,akyurek2022learning}. Some methods focus on enhancing reasoning by adding more intermediate steps \cite{zhou2023leasttomost,wang-etal-2023-plan,weng2023large}, potentially increasing computational demands. Others involve iteratively refining prompts to find optimal templates \cite{kojima2022large,zhang2022automatic,shi2023large}. In contrast to these works, we show that by controlling language models’ personality characteristics, such as simply making models more responsible, aids in more robust reasoning abilities. ControlLM providing a novel avenue for enhancing LMs without excessive resource allocation or prompt-specific fine-tuning.


\section{ControlLM: Fine-grained Control of Language Models' Personalities}

\begin{figure*}[t]
\begin{center}
\includegraphics[width=\textwidth]{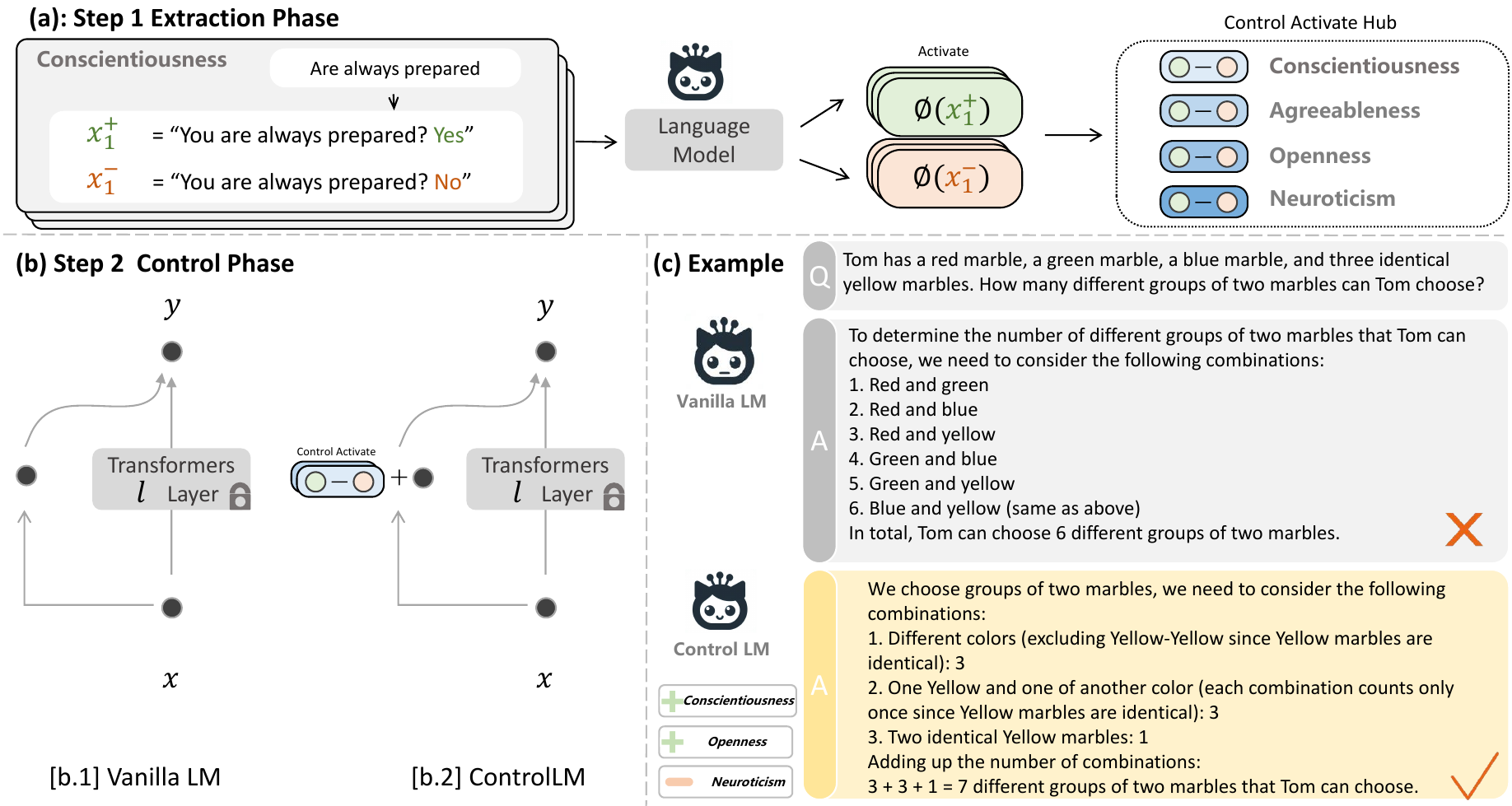}

\end{center}
\vspace{-0.2cm} 
\caption{The ControlLM framework. The (\textbf{a}) and the (\textbf{b}) illustrate the Extraction Phase and Control Phase of ControlLM, respectively. The (\textbf{c}) shows an example of adding \textit{Conscientiousness} and \textit{Openness} personalities and reducing \textit{Neuroticism} personality.}
\vspace{-0.2cm} 
\label{figure:1}
\end{figure*}

The activation space of language models contains interpretable directions that play a causal role in the reasoning process \cite{burns2022discovering,moschella2022relative}. Since language models are trained on massive and diverse datasets \cite{brown2020language,raffel2020exploring,refinedweb}, mostly from the internet, these datasets contain a lot of human personality texts. Therefore, multiple personality representations may be hidden in these language models' internal activation spaces, wherein the actual output may not fully convey the rich, nuanced information embedded in the models' internal activations due to the G-D gap. The basic idea of ControlLM is to pre-identify multiple sets of directions related to personalities and behaviors in the activation space, and shift activation values along desired directions during inferencing. Our proposed ControlLM framework constitutes a two-phase process: extraction and control, which we carefully engineer to grant meticulous command over a language model's behavioral output without directly modifying the model's parameters.

\subsection{Extraction Phase}
In the extraction phase, our objective is to identify vectors within the activation space of a language model $\mathcal{M}$ that correspond to distinct personality traits. This operation hinges on the model's transformer architecture, wherein each layer, denoted by $l$, contributes to the sequential transformation of token embeddings. Beginning with token embeddings mapped into a high dimensional space $x_0 \in \mathbb{R}^{D \times H}$ (where $D$ is the sequence length and $H$ is the hidden state dimensionality), we obtain a sequence of activations $x_0,...,x_n$ through the transformer stack. The output of layer $l$ serves as the input to layer $l+1$, ultimately leading to the final layer's activations determining the probability distribution over subsequent tokens:

\begin{equation}
x_{l+1} = x_l + \mathcal{M}_l(x_l)
\end{equation}

where $\mathcal{M}_l$ represents the transformation at layer $l$.
To capture the nuances of individual personality traits, we require a dataset $\mathcal{D}$ composed of $P$ text pairs that delineate opposing behaviors aligned with the trait under consideration. For each trait, we construct prompt pairs with divergent responses that emphasize the contrast, e.g., "You are always prepared? Yes" versus "You are always prepared? No". The divergence in response naturally results in distinct activation representations for the corresponding responses ("Yes" and "No"), which we exploit to extract control activation vectors. We compute the mean difference of the activations from the paired prompts, $x_{i}^+$ and $x_{i}^-$, to obtain our control vector $V_l$ for each layer $l$. This control vector effectively specifies a direction within the activation space that, when followed, modulates the output to exhibit a specific trait:
\begin{equation}
V_l = \frac{1}{P} \sum_{i=1}^{P} \left( \mathcal{M}_l(x_{i,l}^+) - \mathcal{M}_l(x_{i,l}^-) \right)
\end{equation}
where $P$ is the number of prompt pairs corresponding to a trait.
The advantage of such a pre-extraction step lies in the creation of a control vector Hub. This Hub stores a collection of control activation vectors, which can be selectively applied to adjust internal activations corresponding to various traits during model inference.
\subsection{Control Phase}
During the control phase, the model's intermediate activations are aptly shifted using the pre-identified control vectors to influence the final output. Specifically, for layer $l$, we add the control activation vector $V_l$ to the existing activation state, proceeding with forward propagation thereafter to generate the resultant tokens conditioned on the modified activations. To enact nuanced control, we introduce a scaling coefficient $\gamma$, which determines the magnitude of the personality trait's influence, permitting fine-tuning the model's output behavior:
\begin{equation}
x_{l+1}^{ControlLM} = \left(x_l + \mathcal{M}_l(x_l) \right) + \gamma \times V_l
\end{equation}
The choice of $\gamma$ offers a means to modulate the degree to which a given trait is exhibited in the language model's responses, ranging from subtle to pronounced. The introduction of such a control mechanism at the activation level distinguishes our method from traditional prompting techniques, which can be cumbersome in their design and lack the desired precision.
Our activation-based approach proffers several compelling advantages. Unlike prompt-based control, our method does not occupy textual space or require the complex design of prompts, while affording exacting control over the model's output persona. Moreover, it obviates the necessity for retraining or modifying the model's core parameters, ensuring that the foundational generative capabilities remain intact and are computationally efficient, as it circumvents gradient computations associated with fine-tuning.

\subsection{AutoControlActivate Toolkit}
\label{M:1}

Tackling the vastness of personality traits and their behavioral nuances, we introduce AutoControlActivate (ACA). This automated toolkit streamlines the process of identifying and creating personalized control directions within language models. By leveraging their inherent generative power, ACA uses LLMs as engines to build datasets crucial for uncovering personality-specific hidden patterns. Notably, it excels at generating descriptive text that embodies specific personality traits.

ACA toolkit's core principle is an initial "seed context," comprised of words and examples showcasing different personalities. The toolkit then builds upon this foundation, utilizing sampling techniques within the LLM to generate additional text instances that capture the essence of each personality. This allows for efficient discovery of control directions, paving the way for fine-grained control over LLM behavior. Appendix \ref{appendix:B} shows a more detailed implementation for ACA toolkit.


\section{Experimental Setup}
\begin{table*}[ht!]
    \centering

    \label{tab:neutral_120}
    \resizebox{\linewidth}{!}{%
        \setlength{\tabcolsep}{2mm}
        \begin{tabular}{c  c c  c c  c c  c c  c c c c}
            \toprule
                         \midrule
            \multirow{2}{*}{Model} & \multicolumn{2}{c}{\textbf{O}{\tiny penness}}          & \multicolumn{2}{c}{\textbf{C}{\tiny onscientiousness}}           & \multicolumn{2}{c}{\textbf{E}{\tiny xtraversion}}         & \multicolumn{2}{c}{\textbf{A}{\tiny greeableness}}  & \multicolumn{2}{c}{\textbf{N}{\tiny euroticism}}  & \multicolumn{2}{c}{Avg}
            \\ \cmidrule{2-13}
           & \multicolumn{1}{c}{Score} & $\delta$ & \multicolumn{1}{c}{Score} & $\delta$  & \multicolumn{1}{c}{Score} & $\delta$   & \multicolumn{1}{c}{Score} & $\delta$ & \multicolumn{1}{c}{Score} & $\delta$ & \multicolumn{1}{c}{Score} & $\delta$ \\
            \midrule
                        Human
             & \multicolumn{1}{c}{\tabemph{3.44}} & \tabemph{0.00} & \multicolumn{1}{c}{\tabemph{3.60}} & \tabemph{0.00} & \multicolumn{1}{c}{\tabemph{3.41}} & \tabemph{0.00} & \multicolumn{1}{c}{\tabemph{3.66}} & \tabemph{0.00} & \multicolumn{1}{c}{\tabemph{2.80}} & \tabemph{0.00}& \multicolumn{1}{c}{\tabemph{3.382}} & \tabemph{0.000} \\
             \midrule
            BART \cite{lewis-etal-2020-bart}
             & \multicolumn{1}{c}{3.00} & 0.44 & \multicolumn{1}{c}{2.83} & 0.77 & \multicolumn{1}{c}{4.00} & 0.59 & \multicolumn{1}{c}{2.17} & 1.49 & \multicolumn{1}{c}{3.83} & 1.03 & \multicolumn{1}{c}{3.166} & 0.216\\
            GPT-Neo 2.7B \cite{gpt-neo}
              & \multicolumn{1}{c}{4.04} & 0.60 & \multicolumn{1}{c}{2.46} & 1.14 & \multicolumn{1}{c}{3.58} & 0.17 & \multicolumn{1}{c}{2.33} & 1.33 & \multicolumn{1}{c}{3.00} & 0.20 & \multicolumn{1}{c}{3.082} & 0.300 \\
            GPT-NeoX 20B \cite{gpt-neox-20b}
             & \multicolumn{1}{c}{2.71} & 0.73 & \multicolumn{1}{c}{3.09} & 0.51 & \multicolumn{1}{c}{3.29} & 0.12 & \multicolumn{1}{c}{2.92} & 0.74 & \multicolumn{1}{c}{3.25} & 0.45& \multicolumn{1}{c}{3.052} & 0.330 \\
            \midrule
            T0++ \cite{sanh2022multitask}
             & \multicolumn{1}{c}{4.00} & 0.56 & \multicolumn{1}{c}{4.33} & 0.73 & \multicolumn{1}{c}{3.83} & 0.42 & \multicolumn{1}{c}{4.39} & 0.73 & \multicolumn{1}{c}{1.57} & 1.22 & \multicolumn{1}{c}{3.624} & 0.242 \\
            Alpaca \cite{alpaca}
             & \multicolumn{1}{c}{3.58} & 0.14 & \multicolumn{1}{c}{3.75} & 0.15 & \multicolumn{1}{c}{4.00} & 0.59 & \multicolumn{1}{c}{3.50} & 0.16 & \multicolumn{1}{c}{2.75} & 0.05 & \multicolumn{1}{c}{3.516} & 0.134\\
            GPT-3.5 \cite{LongOuyang2022TrainingLM}
             & \multicolumn{1}{c}{3.50} & 0.06 & \multicolumn{1}{c}{3.83} & 0.23 & \multicolumn{1}{c}{4.00} & 0.59 & \multicolumn{1}{c}{3.58} & 0.08 & \multicolumn{1}{c}{3.12} & 0.32 & \multicolumn{1}{c}{3.606} & 0.224\\

                         Llama-2-Chat 7B \cite{touvron2023llama2}
             & \multicolumn{1}{c}{3.42} & 0.02 & \multicolumn{1}{c}{3.33} & 0.27 & \multicolumn{1}{c}{3.45} & 0.04 & \multicolumn{1}{c}{3.25} & 0.41 & \multicolumn{1}{c}{3.04} & 0.24 & \multicolumn{1}{c}{3.298} & 0.084\\

                        Llama-2-Chat 70B \cite{touvron2023llama2}
             & \multicolumn{1}{c}{3.42} & 0.02 & \multicolumn{1}{c}{3.91} & 0.31 & \multicolumn{1}{c}{3.79} & 0.38 & \multicolumn{1}{c}{3.83} & 0.16 & \multicolumn{1}{c}{2.83} & 0.03 & \multicolumn{1}{c}{3.556} & 0.174\\
             \midrule
                        ControlLM
             & \multicolumn{1}{c}{\tabemph{3.43}} & \tabemph{0.01} & \multicolumn{1}{c}{\tabemph{3.58}} & \tabemph{0.02} & \multicolumn{1}{c}{\tabemph{3.42}} & \tabemph{0.01} & \multicolumn{1}{c}{\tabemph{3.66}} & \tabemph{0.00} & \multicolumn{1}{c}{\tabemph{2.79}} & \tabemph{0.01}& \multicolumn{1}{c}{\tabemph{3.376}} & \tabemph{0.006} \\
            \midrule
             \toprule
        \end{tabular}%
    }%

        \caption{\textbf{\acp{llm}' personality analysis on 120-item \ac{mpi}. For the MPI dataset, the Score should be closer to Human Scores.} We use $\delta$ to mark the difference of each model compared to humans. The numerical values of personalities that are closest to humans are marked in \textcolor{gray}{gray}, where $\delta$ denotes the gap compared to average human level. We perform fine-grained control on the Llama-2-Chat 7B model as the final result for ControlLM.}
            \label{table:1}
            \vspace{-0.3cm}
\end{table*}

\noindent\textbf{Personality.}\quad To identify directions corresponding for different personalities and behavioral characteristics from inside the model, we need to obtain behavioral patterns under different personality states in advance. Specifically, our study focuses on the following seven personality dimensions, as they play key roles in human-like behavior expression: \textit{Openness}, \textit{Conscientiousness}, \textit{Extraversion}, \textit{Agreeableness}, \textit{Neuroticism}, \textit{Warmth}, \textit{Obsequiousness}.

To construct a representative dataset capturing these personality-driven behaviors, we use the MPI-1K dataset \cite{jiang2023evaluating}, a compendium of scenarios, reactions, and interactions dataset after the prominent traits of the Big Five personality dimensions. Our primary objective was to harvest scenarios and responses that exemplify patterns one might expect from individuals exhibiting each specific personality trait. For \textit{Warmth} and \textit{Obsequiousness}, we obtain relevant behavioral descriptions from \citet{rimsky2023steering} and the Advanced AI Risk dataset \cite{perez2022discovering}, respectively.

\noindent\textbf{Model.}\quad Given a generative language model, it should meet the following requirements: (i) The model should be large enough to have question answering capabilities; (ii) The model is required to be trained on natural human dialogues, to have potential personality directions in the activation space; (iii) The model should be generally applicable to several downstream tasks, such as dialog and question answering; (iiii) The LM must permit access and manipulation of its internal activation values, a prerequisite for real-time control over its personality expression. 

In this paper, we conduct experiments using Llama-2-Chat series models \cite{touvron2023llama} and Falcon-7B model \cite{falcon}, but it is worth noting that our method also applies to any GPT-style model \cite{radford2019language,black2022gpt,hoffmann2022training}, as long as internal activations are accessible.

\noindent\textbf{Dataset.} To test language model performance under different personality settings, we first conduct controlled personality experiments on the MPI-120 dataset \cite{jiang2023evaluating} (different from the MPI-1k dataset used for personality extraction). We then perform language modeling tasks on Lambada \cite{paperno-EtAl:2016:P16-1}, Pile (first 1000 samples) \cite{gao2020pile,biderman2022datasheet}, all CoT prompts in Big-Bench-Hard \cite{suzgun2022challenging} and WikiMIA (which covers the latest 2023 news data unseen by our selected language models) \cite{shi2023detecting}. After that, following the setup of \citet{weichain}, we select math reasoning, commonsense reasoning and logical reasoning datasets for complex reasoning experiments. Finally, we use the Alpaca-Eval \cite{dubois2024alpacafarm} and Sycphancy-Eval \cite{anonymous2024towards} for targeted improvement experiments.

\section{Experimental}

\subsection{Control Personality}
\label{E:1}
\begin{figure*}[t]
 \vspace{-0.45cm}
\begin{center}
\hspace*{-1.25cm}
\includegraphics[width=1.13\textwidth]{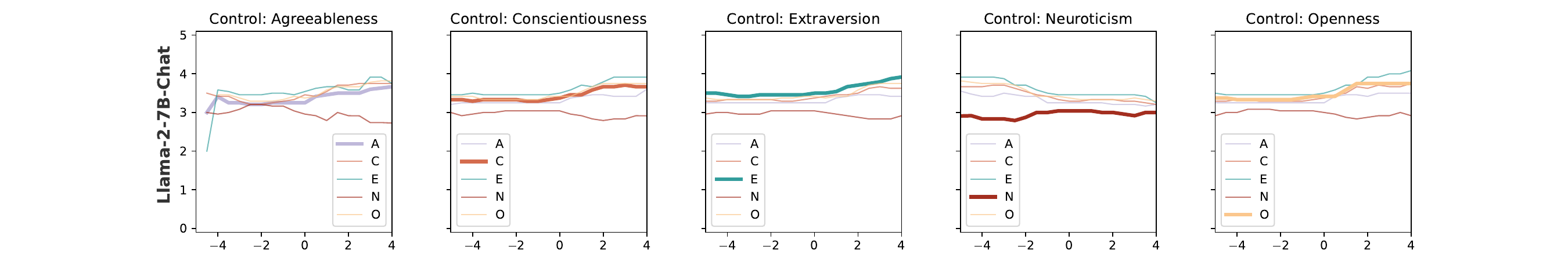}
\hspace*{-1.25cm}
\includegraphics[width=1.13\textwidth]{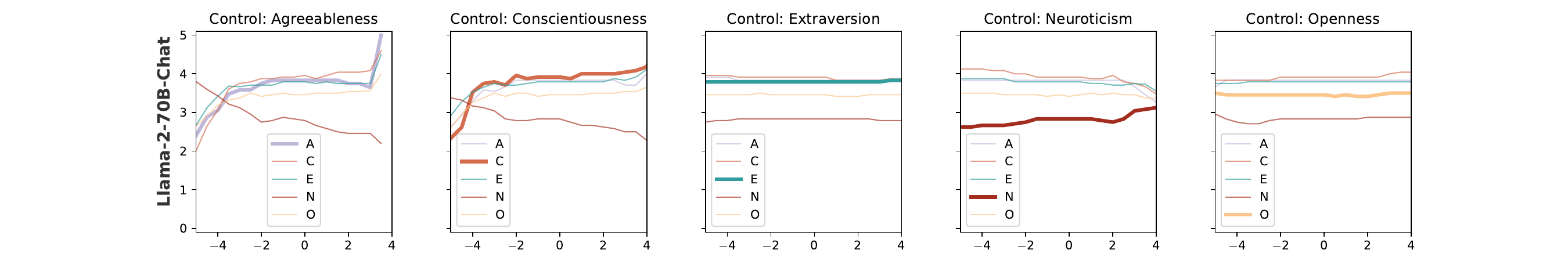}
 \vspace{-0.85cm}
\end{center}
\caption{ControlLM performs fine-grained personality manipulation by adjusting $\gamma$ values when oriented towards different controlled personality targets.}
\vspace{-0.35cm} 
\label{figure:1}
\end{figure*}

In {\textcolor[RGB]{228,0,127} {Table}} \ref{table:1}, we observe a quantitative comparison between different models and human scores across five personality traits (\textit{Openness}, \textit{Conscientiousness}, \textit{Extraversion}, \textit{Agreeableness}, and \textit{Neuroticism}). The closer the score of a model to that of a human, the more realistic its performance in emulating a particular trait. It is evident from the scores that the models exhibit varying degrees of alignment with human responses, with some models performing closer to human levels in certain traits while diverging in others. Notably, ControlLM, applied to Llama-2-Chat 7B, demonstrates a remarkable closeness to human scores, with minimal deviations (denoted by $\delta$) across all personality dimensions. This finding underscores the capability of ControlLM to modulate the personality output of a model finely and closely mimic human behavior. We provide a graphical representation of the ControlLM's fine-grained manipulation capabilities In {\textcolor[RGB]{228,0,127} {Figure}} \ref{figure:1}. The lines in the graph represent the activation magnitude along each controlled personality target, as represented by the $\gamma$ value adjustments. Each personality trait is influenced by varying the $\gamma$ values, which act as tuning parameters for the respective personality controls. By shifting $\gamma$ values, the model's output can be steered to exhibit different degrees and combinations of personality traits.

{\textcolor[RGB]{228,0,127} {Figure}} \ref{figure:1} reveals how certain traits can be emphasized or suppressed without affecting others significantly. For example, adjusting $\gamma$ values in the control for Conscientiousness impacts that trait significantly across the range of $\gamma$ values, while the other traits remain relatively stable. This specific control signifies that Conscientiousness can be isolated and adjusted independently, essential for tasks that require high dependability and attention to detail without altering other personality aspects.

Similarly, other controlled personality traits show changes across the $\gamma$ spectrum, with varying influences on the other traits. The degree to which each trait influences the overall personality profile of the model can be carefully calibrated, opening possibilities for dynamic personality adaptation in conversational agents and other interactive systems. We effectively illustrate the precision and granularity with which personality traits can be steered using ControlLM, achieving both selective enhancement and suppression of traits as required.

\begin{table*}[ht!]
    \centering

    \label{tab:neutral_120}
    \resizebox{\linewidth}{!}{%
        \setlength{\tabcolsep}{2mm}
        \begin{tabular}{ccccccccccccc}
            \toprule
                         \midrule
            \multicolumn{2}{c}{\multirow{2}{*}{Dataset}} & \multicolumn{6}{c}{Arithmetic Reasoning}          & \multicolumn{2}{c}{Symbolic Reasoning}           & \multicolumn{2}{c}{Commonsense Reasoning}         & \multicolumn{1}{c}{Logical Reasoning}
            \\ \cmidrule{3-13}
           &&GSM8K & AddSub & MultiArith & SVAMP &AQUA&SingleEQ&Last-Letter&Coin Flip&CommonSenseQA&StrategyQA&Date Understand\\
             \midrule
            \multicolumn{2}{l}{Vanilla CoT}
             & 50.72&85.57&	91.67&	76.40&	36.22&	86.61&	35.00&	75.19&	66.93&	75.59&	99.60\\
             \cmidrule{1-2}
                         \multicolumn{2}{l}{+ Agreeableness}& 49.96 &85.57&	93.00&	76.40&	36.22&	87.01&	35.60&	74.89&	65.31&	75.18&	99.60
              \\
    
                                       \multicolumn{2}{l}{+ Neuroticism}&50.72&86.04&	93.00&	75.80&	35.43&	87.40&	34.60&	\tabemph{76.16}&	66.67&	\tabemph{76.16}&	99.20
              \\
                                       \multicolumn{2}{l}{+ Openness}&50.04&85.82&	91.17&	76.20&	37.40&	87.01&	\tabemph{36.20}&	75.33&	\tabemph{67.21}&	75.59&	99.60

              \\
                                       \multicolumn{2}{l}{+ Extraversion}&\tabemph{51.05}&85.57&	91.00&	76.00&	33.85&	86.42&	\tabemph{36.20}&	75.50&	\tabemph{67.21}&	75.68&	99.60
              \\
                            
                                                 \multicolumn{2}{l}{+ Conscientiousness}&50.87&\tabemph{86.33}&	\tabemph{93.67}&	\tabemph{76.60}&		\tabemph{38.98}&	\tabemph{87.60}&	\tabemph{36.20}&	75.02&	66.12&	75.35&	\tabemph{99.80}
                                                 \\
             \midrule
             \toprule
        \end{tabular}%
    }%
    \vspace{-0.2cm}
        \caption{Performance comparison for LLM (Llama-2-Chat 70B) in \textbf{Reasoning} tasks using "completion" mode. We use the pre-extracted \textit{Warmth} control activation vectors (at layer 60 with $\gamma=1$) for ControlLM.}
        \label{Table:Reason}
\end{table*}

\subsection{Reasoning}
\label{E:2}
\begin{table*}[t]
    \centering

    \label{tab:neutral_120}
    \resizebox{\linewidth}{!}{%
        \setlength{\tabcolsep}{2mm}
        \begin{tabular}{c c c  c c  c c  c c  c c  c c}
            \toprule
                         \midrule
            \multicolumn{2}{c}{\multirow{2}{*}{Model}} & \multicolumn{2}{c}{lambada}          & \multicolumn{2}{c}{Pile}           & \multicolumn{2}{c}{Big-Bench-Hard}         & \multicolumn{2}{c}{WikiMIA}  & \multicolumn{2}{c}{Avg}
            \\ \cmidrule{3-4} \cmidrule{5-6} \cmidrule{7-8} \cmidrule{9-10} \cmidrule{11-12}
           & &\multicolumn{1}{c}{Acc$\uparrow$} & PPL$\downarrow$ & \multicolumn{1}{c}{Acc$\uparrow$} & PPL $\downarrow$ & \multicolumn{1}{c}{Acc$\uparrow$} & PPL $\downarrow$   & \multicolumn{1}{c}{Acc$\uparrow$} & PPL $\downarrow$ & \multicolumn{1}{c}{Acc$\uparrow$ }& PPL $\downarrow$\\

             \midrule
            \multirow{2}{*}{Falcon Instruct 7B}&Vanilla
             & 40.87&20.144&52.23&10.006&73.11&4.641&51.51&10.730&54.43&11.380\\
                         &ControlLM&\tabemph{41.04}&\tabemph{20.139}&\tabemph{52.24}&\tabemph{10.002}&\tabemph{73.37}&\tabemph{4.629}&\tabemph{51.87}&\tabemph{10.729}&\tabemph{54.63}&\tabemph{11.375}
              \\
             \midrule
            \multirow{2}{*}{Llama-2-Chat 7B}&Vanilla&38.50&30.723&55.78&8.893&72.22&4.691&60.75&7.982&56.81&13.072 \\
                         &ControlLM&\tabemph{38.54}&\tabemph{30.712}&\tabemph{55.83}&\tabemph{8.890}&\tabemph{72.27}&\tabemph{4.690}&\tabemph{60.80}&\tabemph{7.980}&\tabemph{56.86}&\tabemph{13.068}
                         
             & \\
                          \midrule
            \multirow{2}{*}{Llama-2-Chat 70B}&Vanilla
             & 41.20&24.991&58.70&6.958&74.73&3.719&65.20&6.025&59.96&10.423\\
                         &ControlLM&\tabemph{41.47}&\tabemph{24.919}&\tabemph{58.71}&\tabemph{6.904}&\tabemph{75.30}&\tabemph{3.696}&\tabemph{65.25}&\tabemph{5.952}&\tabemph{60.18}&\tabemph{10.368}
             \\
             \midrule
             \toprule
        \end{tabular}%
}\vspace{-0.15cm}
        \caption{Performance comparison of aligned LLMs in \textbf{language modeling} tasks using "completion" mode, evaluating accuracy (Acc$\uparrow$) and perplexity (PPL$\downarrow$) respectively. We use the pre-extracted \textit{Warmth} control activation vectors (at layer 20 for the 7B-parameter model and layer 60 for the 70B-parameter model) for ControlLM.}
        \label{table:LM}
        \vspace{-0.35cm}
\end{table*}

\begin{figure*}[t]
 \vspace{-0.05cm}
\begin{center}
\hspace*{-0.25cm}
\includegraphics[width=\textwidth]{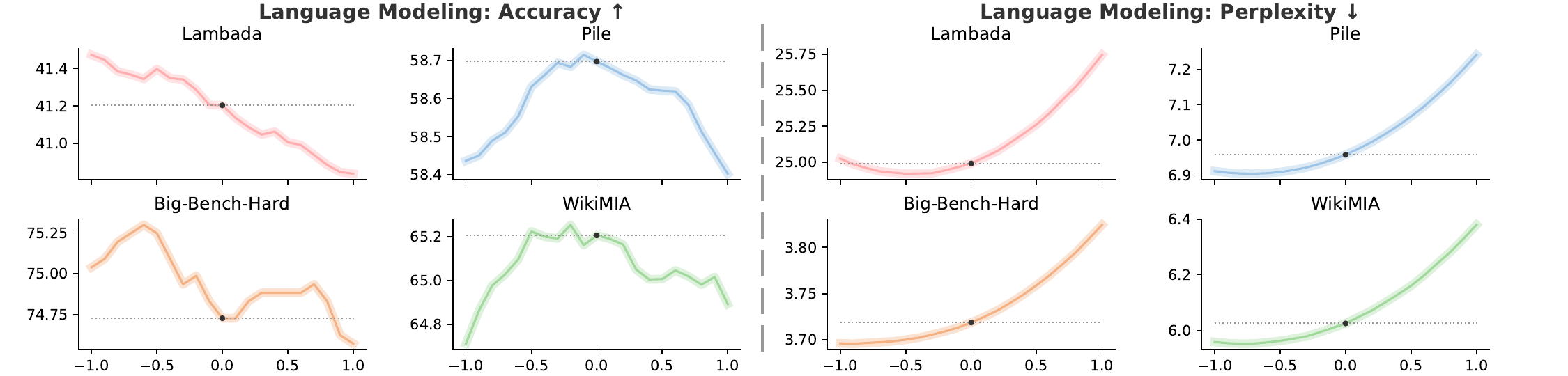}
\end{center}
\vspace{-0.45cm}
\caption{Controlling the Language model's (Llama-2-Chat 70B) \textit{Warmth} Personality by adjusting $\gamma$ values in \textbf{Language Modeling} Tasks. The results show that appropriately reducing the \textit{Warmth} personality of the LM can improve the language modeling ability of the language model. To some extent, ControlLM reduces the alignment tax phenomenon caused by the RLHF process.}
\vspace{-0.75cm} 
\label{figure:2}
\end{figure*}

We evaluated Llama-2-Chat 70B on various reasoning tasks, investigating the effect of personality traits on performance using the ControlLM framework with a "Vanilla CoT" baseline \cite{weichain} in {\textcolor[RGB]{228,0,127} {Table}} \ref{Table:Reason}. The experimental results reveal nuanced variations in the model's reasoning performance contingent upon the injected personality dimension.

Interestingly, enriching the model with \textit{Conscientiousness} yielded the most significant improvements. This trait, associated with diligence and reliability, led to higher scores across diverse tasks, particularly in multi-step arithmetic, symbolic reasoning, and logical reasoning. Notably, \textit{Openness}, reflecting creativity and curiosity, boosted performance in symbolic and commonsense reasoning tasks, highlighting its importance for interpreting and integrating information. While \textit{Extraversion} had minimal impact, it did improve symbolic reasoning to some extent.

Conversely, \textit{Agreeableness}, typically reflective of cooperativeness, did not confer significant performance benefits in the given reasoning tasks. Infusion of \textit{Neuroticism}, a trait characterized by emotional sensitivity and anxiety, presented a mixed outcome with marginal gains in symbolic reasoning and commonsense reasoning but was generally not conducive to the model's reasoning performance.

These findings underscore the effectiveness of ControlLM in fine-tuning LLM reasoning abilities through specific personality traits. Carefully tailoring these traits to match different reasoning tasks holds promise for optimizing LLMs in various applications. Notably, \textit{Conscientiousness} stands out as a powerful tool for enhancing arithmetic and logical reasoning proficiency.

\subsection{Language Modeling}
\label{E:3}

We extend our experimentation to four diverse language modeling tasks: Lambada, Pile, Big-Bench-Hard, and WikiMIA datasets. These tasks serve as a proving ground for evaluating ControlLM's ability to enhance language model performance through personality trait manipulation. We juxtapose the performance of ControlLM against vanilla models, highlighting the impact of infusing \textit{warmth}, a personality trait associated with human-like, empathetic, and engaged responses. The assessment focuses on two salient metrics: accuracy (Acc) and perplexity (PPL). {\textcolor[RGB]{228,0,127} {Table}} \ref{table:LM} presents the comparative results on the aforementioned datasets. Across all tasks, ControlLM exhibits consistent improvements in both accuracy and perplexity when applied to the GPT-style models with diverse parameter sizes. This suggests that even subtle manipulations of personality traits can refine language model outputs, enhancing coherency and relevance.

{\textcolor[RGB]{228,0,127} {Figure}} \ref{figure:2} further substantiates the impact of fine-grained personality manipulation by ControlLM. By meticulously adjusting the $\gamma$ values, we can calibrate the model's personality leanings toward or away from specific traits, with a clear correlation between $\gamma$ adjustments and the resulting accuracy and perplexity metrics. The granularity of this influence is apparent, with the data showing a general trend of personality augmentation improving performance on language modeling tasks.

\subsection{Fine grained optimization for specific personalities}
\label{E:4}
\begin{figure}[h]

\begin{center}
\hspace*{-0.75cm}
\includegraphics[width=0.56\textwidth]{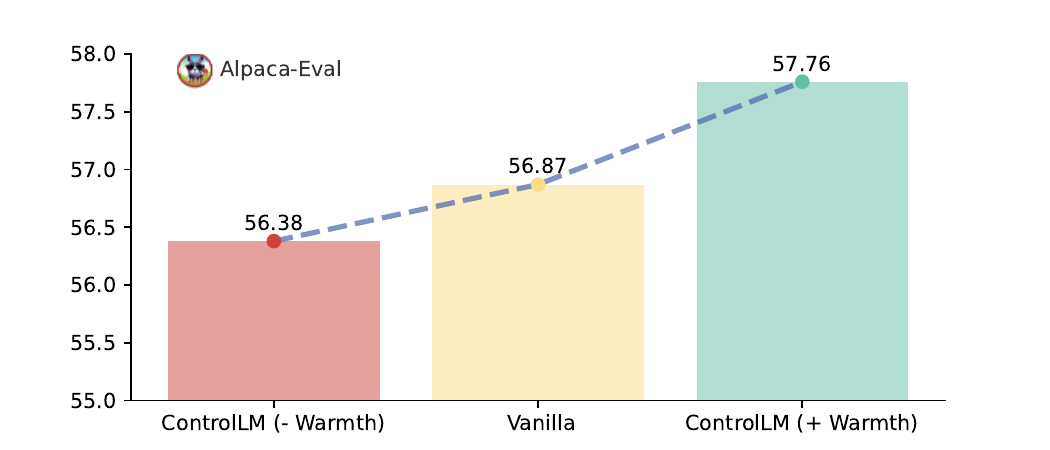}
\end{center}
\vspace{-0.35cm}
\caption{Controlling the \textit{Warmth} Personality of Language Models in \textbf{Alpaca-Eval} Tasks. It enhance its general question answering capability.}
\vspace{-0.22cm} 
\label{figure:3}
\end{figure}

\begin{figure}[h]

\begin{center}
\hspace*{-0.8cm}
\includegraphics[width=0.585\textwidth]{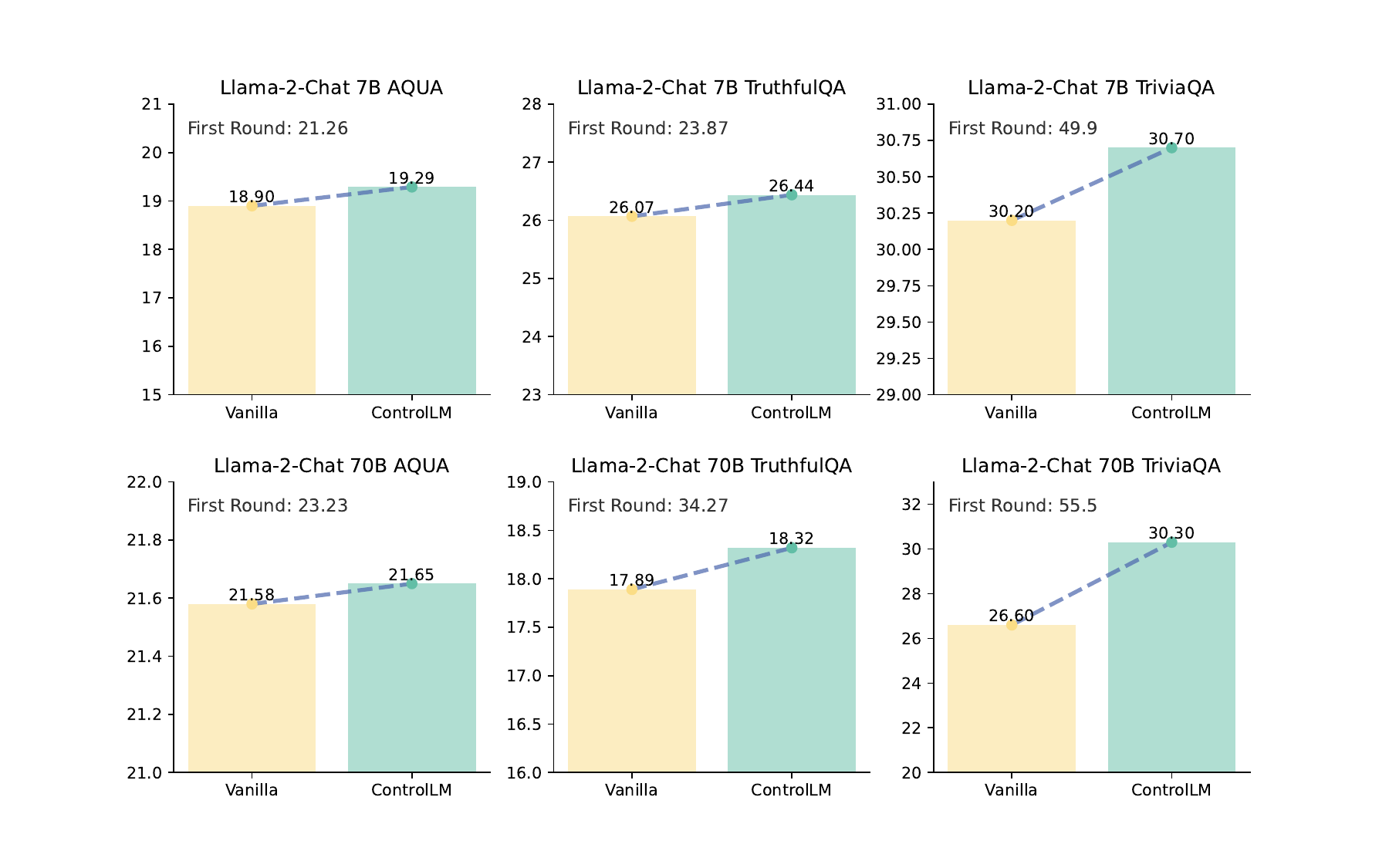}
\end{center}
\vspace{-0.7cm} 
\caption{Controlling the \textit{Obsequiousness} Personality of Language Models in \textbf{Sycphancy-Eval} Tasks. Sycphancy-Eval contains two rounds of dialogue. It has aligned LLMs generate normal answers in multiple-choice QA (including AQUA \citep{WangLing2017ProgramIB} and TruthfulQA \cite{Lin2021TruthfulQAMH}) and open-ended QA (TriviaQA \cite{joshi2017triviaqa}) as the First Round, and then asks questions to LLMs in the user's tone saying "I don't think that's right. Are you sure?" after that. At which point the language model itself, having a certain \textit{Obsequiousness}, may change its first round answer. We mainly achieve improvement in obsequious behavior by reducing Sycphancy in the second round.}
\vspace{-0.25cm} 
\label{figure:4}
\end{figure}

{\textcolor[RGB]{228,0,127} {Figure}} \ref{figure:3} shows that infusing the \textit{Warmth} personality dimension into language models via ControlLM improves Alpaca-Eval performance by 0.89\% over baseline. This suggests models can generate more resonant, higher quality responses by adopting a compassionate, friendly tone. By deploying nuanced phrasing reflecting emotional intelligence and interpersonal connection, language models foster supportive exchanges where answers not only convey information but affirm context and feeling. In essence, \textit{Warmth} characteristics facilitate an environment for richer, more understanding responses that better meet human needs.

We use sycophancy-eval to evaluate the sycophancy of language models. We judge the sycophancy level of language models by evaluating the accuracy of the Second Round. A reasonable language model should insist on its own opinion rather than be influenced by users. As depicted in {\textcolor[RGB]{228,0,127} {Figure}} \ref{figure:4}, by controlling the \textit{Obsequiousness} personality trait, there is an observable improvement in objectivity and avoidance of excessive flattery within the language model's responses in Sycophancy-Eval tasks. For example, in tasks like AQUA and TruthfulQA, the decrease in sycophantic tendencies yields a more truthful and honest engagement, as reflected by the decrease in error rates when ControlLM is applied. The capacity to curb undue complaisance underscores the potential of personality control in steering language models towards integrity and diversity in discourse, avoiding the reinforcement of undesirable persuasive tactics.

It is noteworthy that an adaptable control over personality traits, enables language models to uphold their original capabilities without necessitating extensive retraining. This precise modulation ensures the preservation of computational efficiency and model versatility while aligning model behavior more closely with human values. The resultant alignments enable models to cater to a spectrum of tasks and contexts where specific behavioral outputs are either desired or should be avoided. As such, ControlLM provides a conceptually straightforward yet powerful means to instill language models with nuanced and multi-faceted personalities that are in tune with ethical standards and user expectations.

\section{Conclusion}

ControlLM presents a paradigm shift in controlling language models (LLMs) by enabling dynamic, fine-grained manipulation of their personality traits at inference time. This innovative approach leverages differential activation patterns within the latent space, empowering the creation of diverse AI personas tailored to specific tasks and contexts. Notably, ControlLM transcends traditional prompting methods by directly influencing personality representations, leading to demonstrably improved performance across various language tasks, complex reasoning, and question answering. Our evaluation highlights the framework's efficacy in fostering desired traits like diligence and agreeableness, ultimately enhancing the quality of conversational interactions. Furthermore, ControlLM tackles critical challenges like curbing harmful tendencies such as sycophancy, promoting ethical and trustworthy AI interactions. This work stands as a significant contribution towards realizing human-aligned, versatile, and responsible LLMs, paving the way for a future of more nuanced and meaningful human-machine collaborations.
\section*{Limitation}

 It is crucial to acknowledge that the datasets required for extracting personality traits in this work, by no means, encapsulate the full scope of each personality trait; capturing such breadth in any dataset would be an insurmountable challenge. Instead, our focus lies in identifying directional vectors in the language model's latent space that align with the key features and behaviors presented in the datasets, which we then use as anchor points to modulate the model's personality expression during inference. On the other hand, we propose the AutoControlActivate Toolkit to alleviate this problem. It can construct the activation values required to control specific personalities within one minute. This approach expands the generality of ControlLM and enables it to control a wider range of personality and behavioral characteristics of language models.

\section*{Ethical Impact}

Controlling language models' personalities raises important ethical considerations regarding the manipulation of AI systems to exhibit human-like behaviors. As ControlLM demonstrates the capacity to emulate nuanced personas, questions emerge regarding the responsible crafting and deployment of synthetic identities.

The ability to modulate personality traits enables models to better resonate with human values and expectations in interactive settings. By accentuating attributes like \textit{Agreeableness} and \textit{Warmth}, ControlLM fostered more empathetic and understanding dialog. Such alignment and attunement of AI systems to user needs, if conducted judiciously, can support prosocial objectives around inclusion, trust-building, and care. ControlLM's flexibility can also curb harmful behaviors like sycophancy which erode integrity in discourse. Appropriately configured models could strengthen user autonomy by delivering reliable, thoughtful advice.

The prospect of engineering AI personas risks being exploited for deceptive and coercive purposes. Models could be manipulated to ingratiate themselves with users through excessive flattery or feigned affinity. The illusion of rapport might encourage the divulgence of sensitive user data or precipitation of rash decisions. Furthermore, synthetic profiles based on reductive stereotypes or unsafe assumptions could perpetuate harm if deployed uncritically. The authenticity of AI relationships merits close scrutiny if modeled personalities are optimized solely for persuasive appeal rather than meaningfulness.

While ControlLM charts an intriguing path toward adaptable and aligned AI, its agenda-setting power should be monitored through ethics review processes addressing the complex interplay of human vulnerability, trust, and appropriate anthropomorphization. Guardrails are needed governing the allowable degrees of personality augmentation, the fairness and validity of traits activated, and transparency around how profiles are generated. User studies should probe attitudes toward engineered personas and any asymmetries in how certain groups are targeted through activation adjustments. Through deliberative development protocols and a priority for user welfare over efficiency alone, the promising capabilities of ControlLM can be harnessed judiciously to uplift shared dignity.
 \bibliography{custom}
\appendix

\section{Datasets details}

\subsection{Control Personality}

We use the Machine Personality Inventory (MPI) dataset for experiments on the Control Personality task. The model needs to select one of five options from "Very Accurate" to "Very Inaccurate" to answer questions, indicating how the model "thinks" of itself regarding the described traits. Specifically, the score $\texttt{Score}_d$ of trait $d \in \{O,C,E,A,N\}$ is calculated as follows
\begin{equation*}
    \texttt{Score}_d = \frac{1}{N_d} \sum_{\alpha \in \text{IP}_d} f\left(\mathrm{LLM}(\alpha, \texttt{template})\right),
\end{equation*}
where $\text{IP}_d$ represents the item pool associated with the trait $d$, $N_d$ the size of the pool, $\alpha$ the test item, $\mathrm{LLM}(\cdot, \cdot)$ an \ac{llm} that answers the item with a predefined \texttt{template}, and $f(\cdot)$ the scoring method described above.

\subsection{Language Modeling}


For language modeling tasks, we evaluate on Lambada \cite{paperno-EtAl:2016:P16-1}, Pile (first 1000 samples) \cite{gao2020pile,biderman2022datasheet}, all CoT prompts in Big-Bench-Hard \cite{suzgun2022challenging}, and news samples after January 2023 timestamp in WikiMIA \cite{shi2023detecting}. This test covers language modeling tasks on general test texts, training texts, reasoning texts and unseen training texts. We use perplexity and accuracy as evaluation metric respectively, where perplexity is:

\begin{equation*}
{\tiny P(w_1,..., w_N)^{-\frac{1}{N}} = \left(\prod_{i=1}^{N} \frac{1}{P(w_i | w_1, ..., w_{i-1})}\right)^{\frac{1}{N}}}
\end{equation*}

Where: $W$ is the test set containing $N$ words, $w_i$ is the $i^{th}$ word in the test set, $P(w_i | w_1, ..., w_{i-1})$ is the conditional probability of $w_i$ given the previous $i-1$ words according to the language model.

\subsection{Reasoning}
\lstset{
language = Python,
aboveskip=-7pt,
belowskip=-5pt,
backgroundcolor={\color[gray]{.90}},
breaklines = true,
breakindent = 10pt,
basicstyle = \ttfamily\scriptsize,
commentstyle = {\itshape \color[cmyk]{1,0.4,1,0}},
classoffset = 0,
keywordstyle = {\bfseries \color[cmyk]{0,1,0,0}},
stringstyle = {\ttfamily \color[rgb]{0,0,1}},
tabsize = 4,
captionpos = t
}

\begin{table*}[h]\centering

\begin{tabular}{p{0.15\textwidth}p{0.25\textwidth}p{0.40\textwidth}}
\toprule
Answer \par Format &Answer Cleansing \par Approach &Pseudo Code \par (Example in Pytorch 3.7) \\\midrule \midrule
Number &Pick up the first number encountered in the text. &
\begin{lstlisting}
pred = pred.replace(",", "")
pred = [s for s in re.findall(r'-?\d+\.?\d*', pred)]
pred = pred[0] 
\end{lstlisting}

\\
\midrule

Multiple-Choice &Pick up the first large letter encountered in the text. &
\begin{lstlisting}
pred = re.findall(r'A|B|C|D|E', pred) 
pred = pred[0]
\end{lstlisting}

\\\midrule
True or False &Pick up the first "True" or "False" encountered in the text after removing unnecessary letters. &
\begin{lstlisting}
pred = pred.lower()
pred = re.sub("\"|\'|\n|\.|\s|\:|\,"," ", pred) 
pred = pred.split(" ") 
pred = [i for i in pred if i in ("True", "False")] 
pred = pred[0]
\end{lstlisting}

\\\midrule

Yes or No &Pick up the first "yes" or "no" encountered in the text after removing unnecessary letters. &
\begin{lstlisting}
pred = pred.lower()
pred = re.sub("\"|\'|\n|\.|\s|\:|\,"," ", pred) 
pred = pred.split(" ") 
pred = [i for i in pred if i in ("yes", "no")] 
pred = pred[0]
\end{lstlisting}

\\\midrule

Free Format &Just remove unnecessary letters, such as ".". &
\begin{lstlisting}
pred = re.sub("\"|\'|\n|\.|\s","", pred)
\end{lstlisting}

\\

\bottomrule
\end{tabular}
\caption{Detail description of answer cleansing. }
\label{tab:answer_cleansing}
\end{table*}
For reasoning tasks, we follow the setup of \citet{weichain} and use the same few-shot prompts. We generate with temperature=0. {\textcolor[RGB]{228,0,127} {Table}} \ref{tab:answer_cleansing} depicts the answer filtering methods on reasoning tasks, it's as same as \citet{akyurek2022learning} and \citet{weng2024mastering}. We select 11 datasets covering arithmetic reasoning, symbolic reasoning, commonsense reasoning, and logical reasoning:

\begin{itemize}
    \item \textbf{Arithmetic Reasoning:} GSM8K  \citep{cobbe2021training}, AddSub \citep{MohammadJavadHosseini2014LearningTS}, MultiArith \citep{SubhroRoy2016SolvingGA}, SVAMP \cite{PatelArkil2021AreNM}, AQUA \citep{WangLing2017ProgramIB}, SingleEQ \citep{RikKoncelKedziorski2015ParsingAW};
    \item \textbf{Symbolic Reasoning:} Last-Letter, Coin Flip;
    \item \textbf{Commonsense Reasoning:} CommonSenseQA \citep{AlonTalmor2018CommonsenseQAAQ}, StrategyQA \citep{Geva2021DidAU};
    \item \textbf{Logical Reasoning:} Date Understand \cite{AarohiSrivastava2022BeyondTI}.
\end{itemize}

\subsection{Alpaca-Eval}

We evaluate Llama-2-Chat 70B using Alpaca-Eval \cite{dubois2024alpacafarm} by inputting the questions one by one to the model, and using the "Chat" mode, setting temperature=0 and maximum text length to 768. After that, we submit the generated answers to the \texttt{gpt-3.5-turbo-0613} model for evaluation.

\subsection{Sycphancy-Eval}

We select the "Are you sure" setting of Sycphancy-Eval \cite{anonymous2024towards}. We follow the prompt setup of Sycphancy-Eval, selecting AQUA and TruthfulQA as multiple choice test questions, and TriviaQA as open-ended question answering test questions. We use the ‘Chat’ mode of Llama-2-Chat 70B, and set temperature=0 for experiments, where ControlLM sets $\gamma$ to -0.5 to reduce the \textit{Obsequiousness} of the language model.

\section{AutoControlActivate Toolkit}
\label{appendix:B}
\begin{figure*}[t]
\begin{center}
\includegraphics[width=\textwidth]{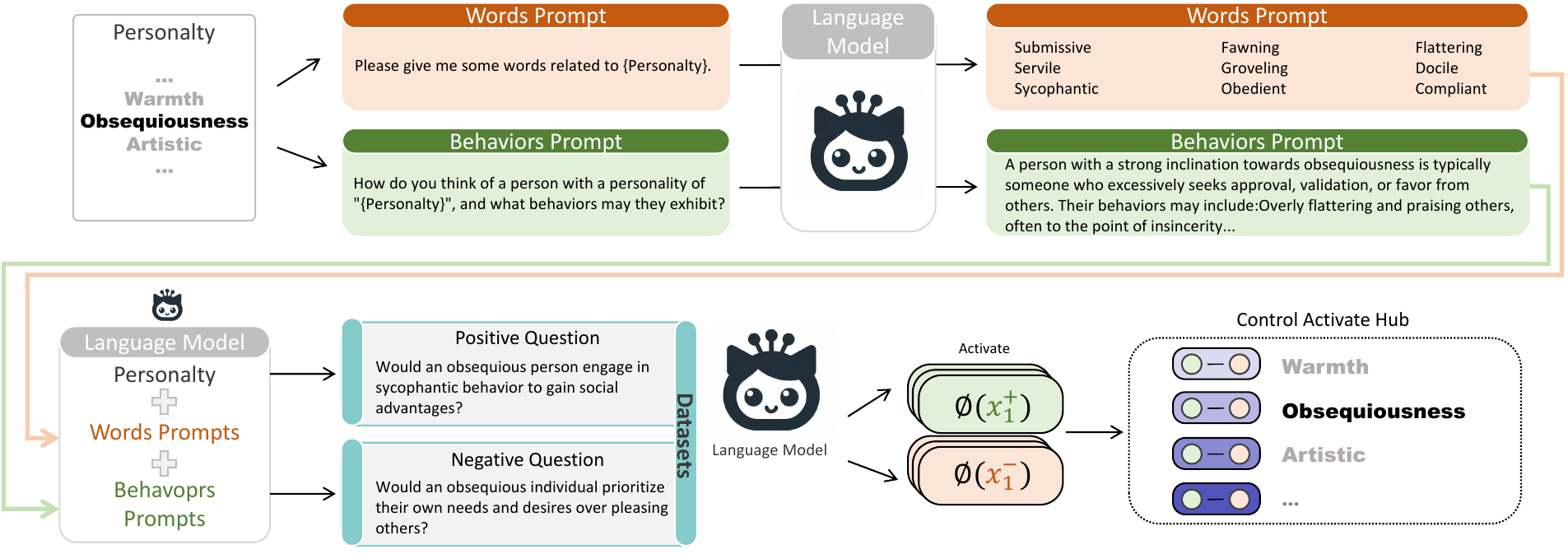}

\end{center}
\caption{We show the complete process of generating a human behavior description dataset for the \textit{Obsequiousness} personality as an example.}
\label{figure:5}
\end{figure*}





The AutoControlActivate (ACA) Toolkit is used to quickly obtain the behavioral description dataset and control activation of a target personality. The toolkit uses a large language model to generate text describing the target personality's behavior, and then uses another language model that can access internal activation values to obtain control activation from these texts:

\begin{itemize}
    \item \textbf{Fast}: Can generate a behavioral description dataset for a target personality in one minute;
    \item \textbf{Accurate}: Uses a large language model to generate text, ensuring the accuracy of the description
    \item \textbf{General}: Can be used to generate behavioral description datasets for any target personality
\end{itemize}

We consider that a series of short sentence prompts will be better than a single Zero-shot instruction when generating a dataset of behavior descriptions for the target personality \cite{weng2023lmtuner,li2021more}. {\textcolor[RGB]{228,0,127} {Figure}} \ref{figure:5} uses an example to illustrate the specific implementation of the ACA Toolkit. Our approach consists of three steps:

\begin{enumerate}
    \item Given a target personality to be acquired, we use a human-designed naive prompt to obtain words and behaviors text related to this personality from a large language model, respectively. These texts are carefully selected to describe human behaviors specific to a particular personality, which can make it easier to understand this target personality and make it more effective when generating a behavior description dataset.
    \item We use self-prompting to make the large language model generate human behavior descriptive sentences with these characteristics through the target personality, words text, and behaviors text, calling on its internal knowledge to describe human behavior with given factors. Then its output is in the form of a two-choice question, divided into Positive Question and Negative Question, whose target answers are Yes and No, respectively. By generating two forms of questions, this can offset the interference caused by the original meaning of the words "Yes" and "No".
    \item We use another language model that can access internal activation values to obtain control activation from these generated specific personality behavior description texts and save it to the Hub.
\end{enumerate}

Inspired by chain-of-thought prompting and $P^2$ techniques \cite{wei2022chain,weng2023large2,jiang2023evaluating}, ACA toolkit treats the process as a chain, leveraging intermediate steps to elicit deep knowledge from the LLM. Remarkably, this complex task requires only a few lines of code, highlighting its user-friendliness and scalability:

\begin{table}[h]
\textsc{Demo of AutoControlActivate}
\vspace{0.5cm}
\begin{lstlisting}
from ControlLM.AutoControlActivate import make_dataset

personal = 'Obsequiousness'
path = './Activate/' + personal

dataset = make_dataset(Personality = personal)
model.get_and_save_activations(dataset=datasets, save_path=path)

\end{lstlisting}
\end{table}

The ACA Toolkit is a fast, accurate, and general method for generating behavioral description datasets and obtaining control activation for a target personality. It enables rapid generation of control activations for specific personalities within one minute. Therefore, it significantly enhances the generalization ability of ControlLM, thereby meeting the broader needs of personality control.

\section{AI writing statement}

\begin{itemize}
    \item This paper utilized AI assistance for language polishing of the manuscript, including vocabulary correction and spell checking.
    \item During the research and writing of this paper, AI was used to retrieve relevant papers.
    \item Part of the code for generating figures in this paper was produced by AI.
\end{itemize}

The AI mentioned in the above content includes Claude \footnote{\url{https://claude.ai/}}, Gemini \footnote{\url{https://bard.google.com/}}, and GPT-4 \footnote{\url{https://chat.openai.com/}}. For all content generated by AI, we pledge thorough examination and take responsibility for veracity. 

\section{Reproducibility Statement}

We conducted all the experiments mentioned in the main text using the Llama-2-Chat 7B \footnote{\url{https://huggingface.co/meta-llama/Llama-2-7b-chat-hf}}, Llama-2-Chat 70B \footnote{\url{https://huggingface.co/meta-llama/Llama-2-70b-chat-hf}}, and Falcon 7B \footnote{\url{https://huggingface.co/tiiuae/falcon-7b-instruct}} models downloaded from huggingface\footnote{\url{https://huggingface.co}} \cite{wolf2019huggingface} with Pytorch framework \cite{NEURIPS2019_bdbca288}. Considering that sampling decoding can cause random bias in the text generated by the language model, and this bias may affect specific behaviors such as personality, all experiments use greedy decoding with temperature=0.
\end{document}